\documentclass[letterpaper, 10 pt, journal, twoside]{IEEEtran}

\usepackage{cite}
\usepackage{microtype}
\usepackage{tikz}
\usepackage{threeparttable}
\usepackage{amssymb}
\usepackage{amsmath}
\usepackage{xcolor} 
\usepackage{booktabs}
\usepackage{graphicx}
\usepackage{mleftright} \mleftright
\usepackage[hidelinks]{hyperref}
\usepackage{titlesec}

\title{Towards a Multi-Embodied Grasping Agent}

\author{Roman Freiberg$^{1}$, Alexander Qualmann$^{1}$, Ngo Anh Vien$^{1}$, and Gerhard Neumann$^{2}$%
\thanks{$^{1}$R. Freiberg, A. Qualmann, and N. A. View are with Bosch Corporate Research.
        {\tt\footnotesize \{roman.freiberg, alexander.qualmann, anhvien.ngo\}@bosch.com}}%
\thanks{$^{2}$G. Neumann is with Autonomous Learning Robots Lab , Karlsruhe Institute of Technology, Germany
        {\tt\footnotesize gerhard.neumann@kit.edu}}%

}

\begin{document}
\bstctlcite{IEEEexample:BSTcontrol}
\maketitle


\begin{abstract}
Multi-embodiment grasping aims to develop approaches that exhibit generalist
behavior across diverse gripper designs. Existing methods often learn the
gripper kinematic structure implicitly and face challenges due to the
difficulty of sourcing the required large-scale data.
In this work, we present a data-efficient, flow-based, equivariant grasp
synthesis architecture that handles different gripper types with variable
degrees of freedom and exploits the underlying kinematic model, deducing all
necessary information solely from gripper and scene geometry.
Unlike previous equivariant grasping methods, we implement all modules
in JAX and provide batching capabilities over scenes, grippers, and grasps,
resulting in smoother learning, improved performance, and faster inference.
Our dataset encompasses grippers ranging from humanoid hands to
parallel-jaw designs, including 25{,}000 scenes and 20 million grasps.
\end{abstract}

\section{Introduction}
\IEEEPARstart{M}ulti-embodiment grasping methods employ a single model to synthesize grasps for
diverse gripper architectures with varying degrees of freedom (DoF). Since
these methods are not tied to one physical system, they adapt effectively to a
broad range of end-effectors. To succeed, approaches must be robust to
hyperparameter choices and hidden biases, enabling the integration of data from
multiple sources~\cite{li2023gendexgrasp, geometrymatching}. Recent work has
combined diffusion models~\cite{julense3, ryu2024diffusion} with architectures
leveraging data symmetries through equivariance~\cite{huorbitgrasp,
gao2024riemann, yang2024equibot, sphericalDP}. These techniques excel at capturing
the multimodal nature of grasping data and currently define the state of the art.
We extend these ideas to the multi-embodiment setting, introducing a single
method capable of synthesizing grasps for grippers with variable DoF.

Our main contributions are: We introduce an equivariant flow-based method for
synthesizing grasps across a wide range of grippers, including humanoid hands
(e.g., Allegro, Shadow Hand) and parallel-jaw grippers (e.g., Franka, ViperX),
encompassing both rotary and prismatic joints. We provide a complete
JAX~\cite{jax2018github} implementation that enables, to our knowledge, for the
first time in this setting, fully batched training and inference across scenes,
grippers, and grasps. This yields smoother learning, faster inference, and lower
memory consumption than prior approaches. Additionally, we release datasets
with joint information for five gripper types and their JAX-integrated kinematic
models, spanning over 25{,}000 scenes, along with data-generation scripts for
both single-object and full-scene tasks. Finally, we demonstrate that our
approach achieves performance on par with the state of the art in both single-
and multi-gripper settings. Code and data are available as open source.\footnote{Code and data available at \href{https://github.com/boschresearch/kinematics-flow}{github.com/boschresearch/kinematics-flow}}

\begin{figure}[!t]
  \centering
  \includegraphics[width=200pt]{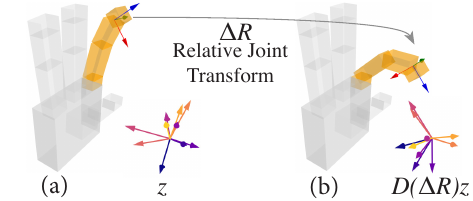}
  
  \caption{
    \textbf{Equivariant Gripper Embeddings.}
    An initial gripper configuration \textbf{(a)} is represented by a learned feature embedding $z$. After a physical joint rotation $\Delta R$, the gripper is in a new configuration \textbf{(b)}. Our method ensures the features are correspondingly transformed via the Wigner-D matrices, $z' = \mathbf{D}(\Delta R)z$, keeping the representation consistent with the physical state.
  }
  \label{fig:equivariant_concept}
\end{figure}

\begin{figure*}[htbp]
  \centering
  \includegraphics[width=\textwidth]{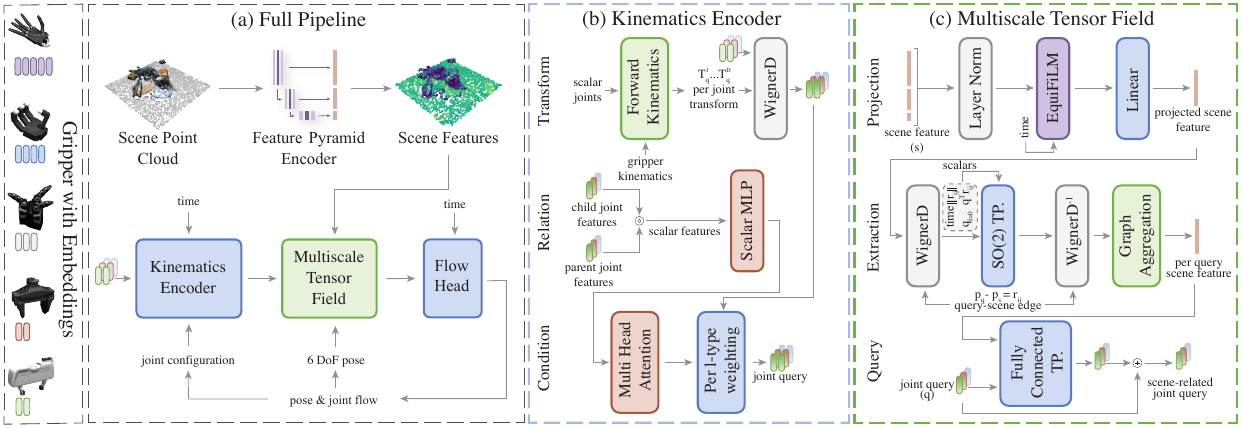}
\caption{\textbf{Method Overview}. 
(\textbf{Left}) Grippers are represented with per-joint equivariant embeddings. 
(\textbf{a}) \textbf{Full Pipeline.} A scene point cloud is encoded into a multi-scale equivariant feature pyramid. Time-conditioned joint features query this pyramid to extract pose and joint information. These scene-aware queries are then decoded to predict flow gradients, which generate the final pre-grasp configuration.
(\textbf{b}) \textbf{Kinematics Encoder.} Joint values and kinematics are used to compute per-joint transformations, which are applied to the embeddings via Wigner-D matrices. Parent-child features interact through a dot product, conditioning the queries with per-$\ell$-type weights.
(\textbf{c}) \textbf{Multiscale Tensor Field.} Hierarchical scene features are time-conditioned using an equivariant FiLM layer \cite{freiberg2025diffusion} and projected to a lower dimension. Relative scene-query positions are encoded via a tensor product dependent on direction and length. The resulting aggregated features for each query are fused with the original joint query via a fully connected tensor product.
}
  \label{fig:main}
\end{figure*}
\section{Related Work}
Grasp detection approaches synthesize grasp poses and configurations, typically categorized into offline methods (generating grasps once) and online approaches (refining continuously during execution). Most research focuses on parallel-jaw grippers~\cite{huorbitgrasp, cgn, breyer2020volumetric, giga}, usually tied to single architectures and utilizing large-scale datasets like ACRONYM~\cite{acronym2020} and GraspNet-1Billion~\cite{billion}.
Dexterous grasping research targets high-DoF architectures. DexGraspNet 2.0~\cite{dex2} learns a diffusion model to predict pre-grasp poses with joint configurations for the Leap gripper in cluttered scenes, while Fast-Grasp’D~\cite{turpin2023fastgraspd} generates multi-finger grasps via differentiable simulation. Dexterous agents employ policies for high-DoF grippers such as the Leap and Shadow Hand; examples include UniDexGrasp++~\cite{uni}, Unified Generative Grasping~\cite{ugg}, and AffordDexGrasp~\cite{wei2025afforddexgrasp}. Furthermore, DexGraspVLA~\cite{zhong2025dexgraspvla} demonstrates effective grasping using pretrained vision-language models, and DexGrasp Anything~\cite{zhong2025dexgrasp} employs a physics-informed diffusion process. MultiGripperGrasp~\cite{casas2024multigrippergrasp}, BoDex~\cite{chen2024bodex}, and Get-a-Grip~\cite{lum2024get} contribute scalable training resources for diverse hands, while GenDexGrasp~\cite{li2023gendexgrasp} demonstrates object-centric features transferring across grippers.
Cross-embodiment representations define contact intent per gripper, treating inverse kinematics as post-processing. Prominent examples include Geometric Matching~\cite{geometrymatching}, learning contact maps per gripper; $D(R,O)$ Grasp~\cite{dro}, relating gripper and object clouds to define contacts; RobotFingerPrint~\cite{fingerprint}, enabling transfer through unified representations; and CEDex~\cite{wu2025cedex}, utilizing human-informed contacts for conditioning. While methods learning gripper-specific encodings~\cite{xu2021adagrasp, zhang2025cross, freiberg2025diffusion, fang2025anydexgrasp} predict stable grasps by incorporating gripper geometry, they do not directly learn all DoFs, relying on fixed configurations~\cite{xu2021adagrasp, freiberg2025diffusion, fang2025anydexgrasp} or eigengrasp reductions~\cite{zhang2025cross}. MADRL~\cite{bonyani2025multi} applies reinforcement learning across multiple grippers but requires specific masking for variable DoF adaptation.

Diffusion- and flow-based generative models learn a transport between an initial uninformative distribution and a target distribution approximated from samples. While methods such as~\cite{diffuser} learn full robot trajectories, work in grasping typically focuses on target end-effector poses on the $\mathrm{SE}(3)$ manifold~\cite{julense3, actionflow, ryu2024diffusion, barad2024graspldm, Chen-RSS-24, lim2024equigraspflow, freiberg2025diffusion}. Protein generative models, which often serve as a source of inspiration, share a similar objective and also operate on the same manifold~\cite{bose2024se3, yim2023se, yim2024improved}.

A surge of large-scale manipulation datasets targets the multi-embodiment problem. RoboMind~\cite{wu2024robomind}, DROID~\cite{khazatsky2024droid}, and AgiBot World Colosseo~\cite{bu2025agibot} collect trajectories for arms and humanoids. Relying on heavy task standardization, these datasets are mutually incompatible.

Generalist agents leverage large-scale data and foundation models to produce policies covering diverse tasks~\cite{kim24openvla, octo_2023, vuong2023open, cross, black2025real}. These methods often struggle with agent-action representations appropriate for multi-embodiment settings. Wang et al.~\cite{wang2025latent} attempt to learn a world model from images, extracting a policy via embodiment-agnostic optical flow fine-tuned for specific embodiments. While demonstrating impressive results, these agents often treat grasping as a downstream task.

Locomotion research mirrors the multi-embodiment trend. Works such as One Policy to Run Them All~\cite{bohlingerone}, Multi-Loco~\cite{yang2025multi}, and Towards Embodiment Scaling Laws in Robot Locomotion~\cite{ai2025towards} demonstrate that increasing morphological diversity improves zero-shot transfer to unknown robots.

Zero-shot domain transfer methods bridge distributions by aligning source and target domains \cite{CycleGAN2017}. In robotics, Mirage~\cite{mirage} and Shadow~\cite{lepert2025shadow} adapt between domains but are limited to single gripper models, introducing biases.

Exploiting symmetries improves robustness and efficiency. Cohen and Welling~\cite{cohen2016group} introduced group-equivariant CNNs for images. Subsequent methods extended this to continuous groups~\cite{harmo, bspline, coord}, including the Euclidean $E(N)$ group, effective in robotics~\cite{huorbitgrasp, lim2024equigraspflow, ryu2024diffusion, gao2024riemann} and imitation learning~\cite{yang2024equibot, actionflow, sphericalDP}. Equivariant layers are often integrated into graph networks~\cite{li2019deepgcns, musaelian2023learning, ryu2024diffusion} to capture local and global structure.

Finally, modern foundation stereo models reduce hardware constraints for point-cloud pipelines. Methods such as FoundationStereo~\cite{wen2025foundationstereo} and the scene reconstruction transformer VGGT~\cite{wang2025vggt} provide reliable depth estimation, reinforcing the feasibility of pure point-cloud models.

\section{Preliminaries}

\subsection{Flows for Distribution Modeling in Robotics}

Continuous Normalizing Flows (CNFs) provide a framework for learning complex probability distributions on smooth manifolds. For robotics applications, we are often interested in the manifold \(\mathcal{M} = \mathrm{SE}(3) \times \mathbb{R}^{D}\). A state is given by \(T=(R, \mathbf{p}, \mathbf{q})\) where \(R \in \mathrm{SO}(3)\) represents orientation, \(\mathbf{p} \in \mathbb{R}^{3}\) position, and \(\mathbf{q} \in \mathbb{R}^{D}\) a general configuration vector.

CNFs learn to transport samples from a simple prior distribution---typically uniform on compact manifolds such as \(\mathrm{SO}(3)\) and normal on open manifolds e.g., \(\mathbb{R}^{3}\)---to a complex target data distribution \(p(T)\). This transport is governed by time-dependent vector fields parameterized by a network with learnable weights $\theta$, defined as
\begin{equation*}
    \boldsymbol\omega_\theta, \mathbf v_\theta: [0,1] \times \mathcal{M} \to \mathbb{R}^{3}, \quad 
    \dot{\mathbf q}_\theta: [0,1] \times \mathcal{M} \to \mathbb{R}^{D},
\end{equation*} where $\boldsymbol \omega, \mathbf v$ and $\dot{\mathbf q}$ are the rotation, position
and joint vector fields, respectively.
These vector fields define velocities, with their outputs residing in the respective tangent spaces \(T_R\mathrm{SO}(3)\), \(T_{\mathbf{p}}\mathbb{R}^{3}\), and \(T_{\mathbf{q}}\mathbb{R}^{D}\). The canonical isomorphism for the tangent space of \(\mathrm{SO}(3)\) is defined via a skew-symmetric matrix basis~\cite{lim2024equigraspflow}. Within the Flow Matching framework, these velocity fields are learned by minimizing the mean-squared error between the predicted flows and ground-truth flows constructed from pairs of samples drawn from the prior and target distributions~\cite{bose2024se3, lim2024equigraspflow, yim2024improved}.

\subsection{Equivariant Representations}
\label{sec:preliminaries}

A function \(f: X \to Y\) is $\mathrm{SE(3)}$-equivariant if transforming its input is equivalent to transforming its output; formally, $f(D_X(r)x) = D_Y(r)f(x)$ for any group action $r \in \mathrm{SE}(3)$ with corresponding representations $D_X$ and $D_Y$. The network layers in this work are designed to be $\mathrm{SO}(3)$-equivariant and translation-invariant. Features are represented as a direct sum of irreducible representations (irreps) of $\mathrm{SO}(3)$, indexed by type $\ell \in \mathbb{N}_0$: $\mathbf{f} = \bigoplus_{\ell=0}^{\ell_{\max}} \mathbf{f}^{\ell}$. Type $\ell=0$ features are scalars, $\ell=1$ are vectors, and higher types capture finer angular detail. A rotation $R \in \mathrm{SO}(3)$ acts on these features via the Wigner-D matrices, $D^{\ell}(R)$.
The primary mechanism for equivariant interaction is the tensor product, denoted by $\otimes$. For two features $\mathbf{x}^{\ell_1}$ and $\mathbf{y}^{\ell_2}$, the output of type $\ell$ is computed via Clebsch-Gordan coefficients (CG Wigner 3j-symbols) as
\begin{equation}
(\mathbf{x} \otimes \mathbf{y})_{\ell m}
=
\sum_{m_1,m_2}
\begin{pmatrix}
  \ell_1 & \ell_2 & \ell \\
  m_1 & m_2 & m
\end{pmatrix}
x_{\ell_1 m_1} y_{\ell_2 m_2}.
\label{eq:weighted-tp}
\end{equation}
We leverage two variants of the tensor product. The \textit{fully connected tensor product} (FCTP) introduces a learnable weight for each valid path in Equation~\ref{eq:weighted-tp}, enabling it to capture rich rotational correlations between features. When feature interactions are conditioned on a specific direction, we can use a more efficient $\mathrm{SO}(2)$-equivariant or sparse channel-wise tensor product. The $\mathrm{SO}(2)$ operation rotates the local coordinate system to align with the direction vector, which simplifies the CG coefficients and renders the operation sparse~\cite{passaro2023reducing}. This technique significantly reduces computational and memory costs and is empirically well-suited for encoding directional information. These operations are foundational to modern architectures like EquiformerV2~\cite{equiformer_v2} or Allegro \cite{musaelian2023learning}.
Geometric and other scalar information is incorporated by modulating the feature vectors. We define a directional modulation operation based on a relative vector $\mathbf{r}_{ji} = \mathbf{r}_j - \mathbf{r}_i$ and a set of arbitrary scalars $s$, formulated as
\begin{equation}
\mathrm{dir\_mod}(\hat{\mathbf r}_{ji}, s)
\;:=\;
\bigoplus_{\ell=0}^{\ell_{\max}} \mathrm{MLP}_\ell( \mathrm{enc}(s) ) \; Y^{\ell}(\hat{\mathbf r}_{ji}),
\label{eq:dir-mod}
\end{equation}
where $Y^{\ell}$ are the real spherical harmonics. The function $\mathrm{enc}(\cdot)$ encodes the input scalars (e.g., distance, time) using suitable bases, and a multi-layer perceptron (MLP) then produces a set of weights that modulate the different irrep types $\ell$. This is analogous to the gating mechanism in Allegro~\cite{musaelian2023learning}. This modulation is critical to our approach, as it provides the mechanism to condition high-degree equivariant features on one-dimensional inputs such as joint angles.

\section{Problem Formulation}
We consider an arbitrary gripper $g$ with an internal joint configuration \(\mathbf q_g \in \mathbb{R}^{D_g}\), where \(D_g\) denotes its DoF, such as \(D_g=2\) for a parallel-jaw gripper or \(D_g=22\) for a dexterous hand. The gripper is modeled as a three-dimensional manifold \(\mathcal{M}_{g}\) with known geometry. Its forward kinematics are described by the function
\(
\mathrm{kin}_g\colon \mathbb R^{D_g} \times \{1, \dots, D_g\} \to \mathrm{SE}(3),
\)
which maps a joint configuration $(\mathbf q_g, i)$ to the transform $T_i~=~(R_i, \mathbf p_i)$ of the $i$-th joint relative to the gripper's base frame.
A scene is composed of graspable objects and potential obstacles, observed as a point cloud \(\mathbf o_s\) in a global coordinate frame. The objective is to learn the conditional 
distribution defined through
\(
p(T_g, \mathbf q_g \mid \mathcal{M}_g, \mathbf o_s),
\)
where the model leverages the known kinematics $\mathrm{kin}_g$. Here, \(T_g \in \mathrm{SE}(3)\) represents the desired pre-grasp pose of the gripper in the scene frame, and \(\mathbf q_g \in \mathbb{R}^{D_g}\) specifies the corresponding pre-grasp joint configuration. A grasp, defined by the pair $(T_g, \mathbf q_g)$, is successful if, after positioning the gripper at this pre-grasp state, a subsequent pre-defined closing motion results in a stable configuration that ensures secure object manipulation.

\begin{figure*}[htbp]
  \centering

  \tikz[remember picture] \node[anchor=center, inner sep=0] (imagebox) {
      \includegraphics[width=\textwidth]{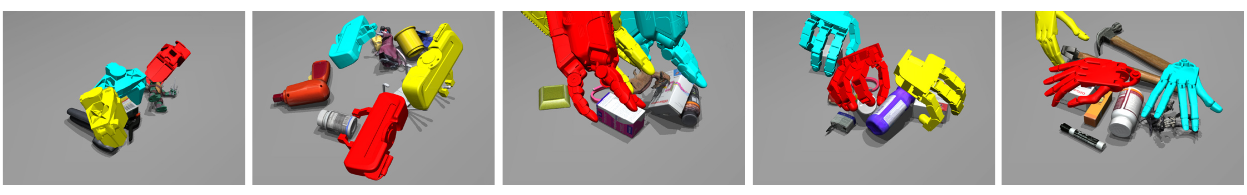}
  };

  \begin{tikzpicture}[remember picture, overlay]
      \begin{scope}[font=\small, text=black] 
          \node at ([xshift=2.0cm, yshift=-3.cm]imagebox.north west) {(a) ViperX};
          \node at ([xshift=5.5cm, yshift=-3.cm]imagebox.north west) {(b) Franka};
          \node at ([xshift=9.1cm, yshift=-3.0cm]imagebox.north west) {(c) DEX-EE};
          \node at ([xshift=12.8cm, yshift=-3.0cm]imagebox.north west) {(d) Allegro};
          \node at ([xshift=16.45cm, yshift=-3.0cm]imagebox.north west) {(e) Shadow};
      \end{scope}
  \end{tikzpicture}

  \vspace{-3mm}

  \caption{\textbf{Multi-Embodiment Grasp Synthesis Examples.} Renderings of three sampled pre-grasp configurations for 
  five distinct grippers in cluttered scenes. Included grippers \textbf{(a)}~ViperX~300s parallel gripper, \textbf{(b)}~Franka Emika parallel gripper, \textbf{(c)}~DEX-EE dexterous hand, \textbf{(d)}~Allegro Hand, and \textbf{(e)}~Shadow Hand.}
  \label{fig:grippers}
\end{figure*}
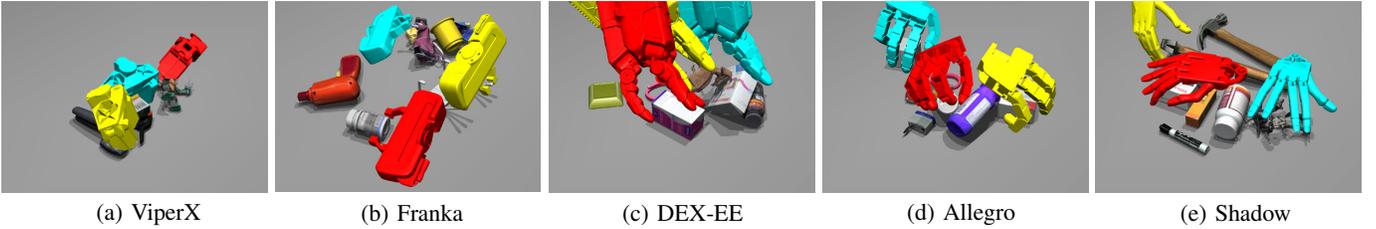

\section{Method}
Our pipeline begins with a geometric scene encoder that converts
raw point-cloud observations into a feature map representing the geometry. For each
time step, the kinematics encoder encodes the gripper configuration only through
transform information of the forward kinematics defining joint queries.
In the Multiscale Tensor Field, joint queries extract grasp-relevant scene information
and relate the relative pose of the gripper and scene. In the Flow head, pose and joint
gradients are decoded for the flow process.
Figure~\ref{fig:main} provides an overview of our pipeline.
Our method is inspired by architectures such as~\cite{huorbitgrasp, ryu2024diffusion,
musaelian2023learning} but is rebuilt from the ground up in the JAX~\cite{jax2018github}
framework, with parallelization and memory efficiency as the main decision drivers.
The resulting architecture allows batching over all grippers, multiple scenes,
and grasps in both single- and multi-GPU training schemes.

\subsection{Geometric Scene Encoder}
\label{sub:geo_enc}

The scene geometry is represented by a point cloud $\{\mathbf r_i\}_{i=1}^N$, typically obtained from dense RGB-D. While color could be included as a scalar feature, our method relies exclusively on geometric information, avoiding the challenges of photorealistic rendering and domain randomization required to align simulated RGB data with real-world conditions. The raw point cloud is voxel down-sampled (1\,mm grid) and mapped to a fixed cardinality of 15,000 points using Farthest-Point Sampling (FPS). This fixed size accommodates JAX's~\cite{jax2018github} static array requirement.

Our encoder constructs a hierarchical representation through a feature pyramid of $L$ stages. At the initial stage, we compute an equivariant feature encoding $\mathbf{f}_i^{(0)}$ for each point $\mathbf{r}_i$ by aggregating information from its local neighborhood $\mathcal{N}(i)$ as a weighted sum
\(
\mathbf{f}_i^{(0)} = \sum_{j \in \mathcal N(i)} w_{ij} \ \mathrm{dir\_mod}\bigl(\hat{\mathbf{r}}_{ji}, r_{ji}\bigr).
\)
Here, $\mathrm{dir\_mod}$ encodes the relative direction $\hat{\mathbf{r}}_{ji}$ and distance $r_{ji}$ from a neighbor $\mathbf{r}_j$ to the center $\mathbf{r}_i$. The weights $w_{ij}$ are distance-based attention scores, learned via $w_{ij} = \mathrm{softmax}_j(\mathrm{MLP}(r_{ji}))$.

Each subsequent pooling stage $k \in \{1, \dots, L\}$ downsamples the cloud $\{(\mathbf r_i^{(k-1)}, \mathbf f_i^{(k-1)})\}$ using FPS to produce a sparser set $\{\mathbf r_i^{(k)}\}$. Features for these points are computed by aggregating from the previous denser layer using the sparse directional $\textrm{SO(2)}$ tensor product in EquiformerV2 modules. This creates a multi-scale representation $\{(\mathbf r_i^{(k)}, \mathbf f_i^{(k)})\}_{k=0..L}$. To expand the receptive field, we apply multiple equivariant message-passing layers with K-Nearest Neighbors (K-NN) graphs. This hierarchical structure enables efficient, point-wise feature querying by concatenating features across all stages. Inspired by Sonata~\cite{wu2025sonata}, which addresses the geometric shortcut problem in U-Nets, we eliminate the decoder path. This significantly reduces memory utilization compared to similar equivariant models~\cite{sphericalDP, ryu2024diffusion}, while preserving a rich, multi-scale representation.

\subsection{Geometric Gripper Encoding}

The gripper encoder produces configuration-aware, equivariant query features representing the gripper's state. A naive encoding of scalar joint values \(\mathbf q_g\) would force implicit learning of kinematics, hindering generalization. We adopt an explicit, geometry-aware approach.

For each joint $i$, we introduce a unique, learnable, equivariant feature embedding $\mathbf{z}_i$, composed of multiple irreducible representations (irreps), e.g., $\ell_0 \oplus \ell_1 \oplus \ell_2$. At minimum, we use scalar ($\ell=0$) and vector ($\ell=1$) components. The forward kinematics $\mathrm{kin}_g(\mathbf q_g, i) = T_i = (R_i, \mathbf p_i)$ place and orient these features in the gripper's base frame. Specifically, rotation $R_i$ is applied to its embedding $\mathbf{z}_i$ via Wigner-D matrices: $\mathbf{z}'_i = \bigoplus_{\ell} D^{\ell}(R_i) \mathbf{z}_i^\ell$.
This transformation ensures the representation covaries with the physical state (visualized in Figure~\ref{fig:equivariant_concept}), aligning the embedding with the joint's local orientation.
To recover scalar features of a revolute joint, we leverage the kinematic chain. We compute a channel-wise dot product between the rotated child embedding $\mathbf{z}'_i$ and its parent's rotated embedding, producing scalars as a function of the relative rotation. These scalars modulate the channels of $\mathbf{z}'_i$ via a lightweight attention mechanism and the $\textrm{dir\_mod}$ operator, conditioning it on the gripper configuration. This mechanism only requires the relative transform between parent and child, making it robust to coordinate system choices. For root joints, a static auxiliary embedding serves as an anchor and is discarded afterwards. To incorporate prismatic joints, we apply an equivariant message-passing layer over the kinematic graph. Messages utilize $\textrm{dir\_mod}$ on relative positions $(\mathbf{p}_i - \mathbf{p}_j)$, encoding the translational component.

Finally, to represent the gripper's 6-DoF global pose, we introduce two additional learnable embeddings anchored at the gripper origin. In the decoding phase, the $\ell=1$ components of these predict the rotational and translational velocity fields (the twist). Crucially, this circumvents canonical coordinate frame selection challenges present in works such as DiffusionEDFs~\cite{ryu2024diffusion} and~\cite{freiberg2025diffusion}. Local frames of different grippers need not be manually pre-aligned; kinematic origins suffice. The learnable embeddings allow the model to implicitly establish a consistent internal frame. This yields $D_g+2$ query features, each with an associated position, explicitly encoding the gripper structure. This flexibility makes the equivariant design a fundamental requirement for our multi-embodiment approach.

\subsection{Gripper-Scene Relation}

The Multiscale Tensor Field module relates gripper query features to encoded scene geometry, providing context for predicting gripper pose and adapting finger placement. First, the $D_g+2$ gripper query features and their positions are transformed from the local frame to the global scene frame based on the current SE(3) pose estimate.

We construct a bipartite graph to establish neighborhoods. For each query, we identify K-nearest neighbors among scene points at the highest-resolution pyramid stage. Leveraging the many-to-one mapping from encoding, we retrieve corresponding features from all hierarchy levels. These hierarchical features are projected to a lower-dimensional latent space and conditioned on flow-time $t$ using an equivariant FiLM layer~\cite{freiberg2025diffusion}.

For each edge connecting a query to a scene point, we apply the edge-direction-dependent $\mathrm{SO}(2)$ tensor product~\cite{passaro2023reducing} on the scene feature. This is conditioned on scalars derived from flow-time, relative distance, and dot products of query and scene features. Resulting edge features are aggregated into a scene-context feature using attention, then fused with the original gripper query using a final FCTP. Note that expensive FCTP operations are only applied per query, limiting computational impact. This block can be stacked to build detailed representations. Finally, scene-aware query features transform back into the gripper's local frame for decoding.

\subsection{Flow Decoding}

After enriching gripper queries with scene context, we apply additional equivariant message-passing layers over the kinematic graph. The final flow is decoded by extracting specific components: $\ell=1$ components of the two pose tokens yield rotational velocity \(\boldsymbol\omega\) and translational velocity \(\mathbf v\), while $\ell=0$ components of the \(D_g\) joint tokens yield joint-space velocities \(\dot{\mathbf q}\). The model is trained by minimizing a weighted mean squared error between predicted flow \((\boldsymbol\omega, \mathbf v, \dot{\mathbf q})\) and ground-truth samples, following common flow-matching frameworks~\cite{lim2024equigraspflow, yim2023se, bose2024se3, yim2024improved}.

\begin{table}[h]
\centering
\caption{Parallel-jaw grasp performance: Success and declutter rate}
\label{tab:orbit}
\begin{tabular}{lcc}
\toprule
\textbf{Method} & \textbf{Success Rate (\%)} & \textbf{Declutter Rate (\%)} \\
\midrule
OrbitGrasp \cite{huorbitgrasp} & \textbf{98.4} & \textbf{98.2} \\
Ours       & {93.7} & 96.2 \\
\bottomrule
\end{tabular}

\begin{minipage}{\columnwidth}
    \vspace{2mm} 
    \footnotesize 
    Grasp success and declutter rates on the OrbitGrasp \cite{huorbitgrasp} benchmark over 100 scenes.
    Models are trained on 2\,500 scenes. While competitive, our method 
    uses less than half the grasps during training and applies no post-processing.
\end{minipage}
\end{table}

\section{Experiments}

We compare our method with the state-of-the-art grasping method OrbitGrasp \cite{huorbitgrasp} and in multi-embodiment grasping settings. It is worth noting that, compared with the previously mentioned methods, ours does not prioritize grasp success rate but rather reproducing the distribution of successful grasps. This distinction provides flexibility in adapting our method to the desired downstream task with minimal modifications.
We validate our approach in a real-world grasping setting.

\subsection{Single Embodiment Parallel Jaw Grasping}
First, we evaluate the feasibility of our approach on
the benchmark presented in OrbitGrasp~\cite{huorbitgrasp} for $\textrm{SE(3)}$ 
grasp pose prediction of the Panda gripper in cluttered object scenes.
For a direct comparison with OrbitGrasp~\cite{huorbitgrasp}, we reproduced its original dataset, which comprises 2,500 cluttered `pile' scenes containing approximately 2.5~million successful and 3.5~million unsuccessful grasps. Adhering to the original evaluation protocol, we assess the grasp success rate and declutter rate in a simulated decluttering procedure across 100 scenes. The results are presented in Table~\ref{tab:orbit}. For the OrbitGrasp model, we select the variant with irreps up to $\ell=2$, as this aligns with our model's architecture and has near-negligible performance differences to significantly more computationally expensive versions with higher $\ell-$types~\cite{huorbitgrasp}.

The results show that while OrbitGrasp~\cite{huorbitgrasp} nearly saturates the success rate metric, our method remains competitive. Notably, our model achieves this performance despite being trained exclusively on successful grasps---utilizing less than half of the available grasp data---and without requiring scene normal or color information as input.
Furthermore, there are two key methodological differences. First, our approach processes the full, high-resolution scene point cloud (15,000 points) at once, in contrast to OrbitGrasp which operates on multiple low-resolution scene patches of fewer than 1,000 points each. Second, the OrbitGrasp architecture is constrained to predicting per-point grasp scores along the object's surface normal. This inherent design choice limits its applicability to more general datasets, such as ours, where grasp poses can be freely positioned within the scene.

\begin{table}[h]
\centering

\caption{Dataset composition by gripper type}
\label{tab:dataset_composition}

\setlength{\tabcolsep}{3pt} 

\begin{tabular}{ccccc}
\toprule
\textbf{Franka} & \textbf{ViperX} & \textbf{DEX-EE} & \textbf{Allegro} & \textbf{Shadow} \\
\midrule
6.87M & 6.24M & 1.58M & 2.12M & 2.99M \\
\bottomrule
\end{tabular}

\begin{minipage}{\columnwidth}
    \vspace{2mm} 
    \footnotesize 
    Total available grasps (in millions) per gripper over all scenes.
    For each gripper we generate 5{,}000 scenes and re-evaluate
    all stable gravity-free grasps per object in the scene.
\end{minipage}

\end{table}

\subsection{Dataset Generation}

\begin{table}[h]
\centering

\caption{Grasp success rate comparison for static grippers}
\label{tab:static_gripper_comparison}

\begin{tabular}{lccc}
\toprule
\textbf{Method} & \textbf{Franka} & \textbf{DEX-EE} & \textbf{Shadow} \\
\midrule
Diffusion \cite{freiberg2025diffusion}  & 97.1 & \textbf{93.5} & 85.8 \\
Ours       & \textbf{97.7} & 84.8 & \textbf{91.0} \\
\bottomrule
\end{tabular}

\begin{minipage}{\columnwidth}
    \vspace{2mm}
    \footnotesize
    Grasp success rate comparison of methods trained on static grippers,
    following Diffusion~\cite{freiberg2025diffusion}, for a direct comparison.
    We only predict flow and diffusion for the pose and fix the joint configurations.
\end{minipage}

\end{table}

Our evaluation requires a dataset composed of scenes containing grasps with explicit joint configurations for each gripper type. The data generation pipeline from Diffusion~\cite{freiberg2025diffusion} is unsuitable for our purposes, as it only provides grasps in fixed open and closed configurations and employs the Antipodal~\cite{billion} method, which is not designed for generating diverse, high-DoF grasps. For our dataset, we follow a two-stage process. First, we generate stable grasps for each gripper-object pair in a gravity-free environment. Second, these candidate grasps are transferred to cluttered scenes, where they are filtered for collisions and their stability is re-evaluated via a lifting motion, similar to the original protocol~\cite{freiberg2025diffusion}.

For the Panda and ViperX parallel-jaw grippers, we employ the Antipodal~\cite{billion} grasp generation method to collect 5,000 grasps per object. For the dexterous grippers (DEX-EE, Allegro, and Shadow Hand), we leverage a differentiable kinematics model implemented in JAX~\cite{jax2018github}, identical to the one used in our main method. We define a set of potential contact points on each gripper's fingertips. For each object, we first sample 10\,cm grasping regions by applying FPS to the object's surface points. Within each region, we randomly select target contact points on the object and corresponding target points on the gripper's fingers; object contact points are slightly offset from the surface to mitigate initial collisions. We then solve for the corresponding joint configurations via gradient descent by minimizing an objective function. The loss terms consist of a positional consistency loss and an angular deviation between surface normals on the fingertips.
Using this single parallelizable method, we can synthesize a large quantity of grasps in a single batch for the used dexterous grippers.
Grasp stability is validated in the MuJoCo~\cite{todorov2012mujoco} physics simulator. A candidate grasp is first checked for collisions. If collision-free, the gripper is closed, and a series of impulse forces are applied to the object. A grasp is deemed stable if the gripper maintains contact throughout all checks. This impulse-based procedure is significantly faster than the shaking-motion approach used in~\cite{freiberg2025diffusion}, while empirically showing more stable grasps. Finally, the validated stable grasps are placed back into the cluttered scenes, following the final data processing steps of~\cite{freiberg2025diffusion}. Table~\ref{tab:dataset_composition} summarizes the total number of grasps generated for each gripper. Due to their geometric complexity and higher degrees of freedom, the dexterous grippers consequently have a lower final grasp density per scene in our dataset.
In total, our generated dataset encompasses 25{,}000 scenes with over 200 objects sourced from the
Google Scanned Objects \cite{downs2022google} and YCB \cite{calli2015ycb} datasets. Each scene contains one to seven objects and
includes up to 1{,}500 grasps.

\begin{table*}[t]
\centering

\caption{Multi-embodiment grasp success rates, coverage, and joint variance.}
\label{tab:multi_embodiment_success}

\setlength{\tabcolsep}{4pt}
\begin{tabular}{l c c c c c c c c c c}
\toprule

& \multicolumn{2}{c}{\textbf{Franka (2 DoF)}} 
& \multicolumn{2}{c}{\textbf{ViperX (2 DoF)}} 
& \multicolumn{2}{c}{\textbf{DEX-EE (12 DoF)}} 
& \multicolumn{2}{c}{\textbf{Allegro (16 DoF)}} 
& \multicolumn{2}{c}{\textbf{Shadow (22 DoF)}} \\

\cmidrule(lr){2-3} 
\cmidrule(lr){4-5} 
\cmidrule(lr){6-7} 
\cmidrule(lr){8-9} 
\cmidrule(lr){10-11}

\textbf{Method} & 
\textbf{Succ.} & \textbf{NJD.} & 
\textbf{Succ.} & \textbf{NJD.} & 
\textbf{Succ.} & \textbf{NJD.} & 
\textbf{Succ.} & \textbf{NJD.} & 
\textbf{Succ.} & \textbf{NJD.} \\
\midrule

Single-Embodiment & 
\textbf{98.1} & \textbf{0.249} & 
\textbf{98.3} & 0.148 & 
\textbf{80.4} & {0.110} & 
82.9 & 0.281 & 
79.3 & 0.218 \\

Multi-Embodiment & 
97.9 & 0.247 {\scriptsize (\textit{-0.8\%})} & 
95.9 & \textbf{0.172} {\scriptsize (\textit{+16.2\%})}& 
74.4 & \textbf{0.115} {\scriptsize (\textit{+4.5\%})}& 
\textbf{86.2} & \textbf{0.295} {\scriptsize (\textit{+5.0\%})}& 
\textbf{80.7} & \textbf{0.252} {\scriptsize (\textit{+15.6\%})} \\

\bottomrule
\end{tabular}

\begin{minipage}{\textwidth}
    \vspace{2mm}
    \footnotesize 
    Grasp success percentage (\textbf{Succ.}) and mean normalized joint deviation (\textbf{NJD.}) per gripper evaluated over 10 scenes testing 100 sampled grasps. 
    Training dataset for each gripper encompasses 5{,}000 scenes.
\end{minipage}

\end{table*}

\subsection{Multi-Embodiment Grasping}

For each gripper, we measure the grasp success rate and joint diversity across 10 held-out test scenes, executing 100 grasps sampled from our model in each scene. During pre-processing, we filter out any grasps that are in collision with the scene. Our optimized implementation can sample a full batch of 100 collision-free grasps for the dexterous grippers in under 10~seconds on a single consumer-grade GPU ($<$12GB~VRAM).

For the single-embodiment benchmark, we train a dedicated model for each gripper type on a single H200 GPU for up to 500 scene-epochs. We use a batch size of five scenes, from which we sample 128 grasps per batch. Note that since scenes can contain up to 1,500 valid grasps, each epoch consists of a different random subset. We employ several data augmentation techniques: randomizing the Farthest Point Sampling (FPS) initialization, applying random rotations around the z-axis, and over-sample later time steps in the flow, which are critical for refining the final pose. As noted in~\cite{huorbitgrasp}, rotational augmentations are beneficial even for equivariant models to mitigate the accumulation of minor numerical errors across layers.

For the multi-embodiment task, we train a single, unified model with a same-sized scene batch count distributed across all five gripper types. To improve conditioning, we incorporate a 
technique from Lim et al.~\cite{lim2024equigraspflow} by adding a 
sixth `dummy' gripper with zero DoFs. This dummy gripper serves as an unconditional class, which 
helps steer the flow generation process more effectively for the specific, conditioned gripper 
types. During training, a fraction of the data from all grippers is repurposed for this dummy 
class by replacing the true gripper label and nullifying the joint information.
To compute a standard diversity metric that is applicable for all gripper types including
parallel jaw, we deviate from the common angle variance metric \cite{dro} in favor of the mean 
normalized joint deviation (NJD). This metric normalizes all joint spaces to between plus and 
minus one and computes the mean of the standard deviations over all DoFs.
The results are presented in Table~\ref{tab:multi_embodiment_success}.
The multi-embodiment results are on par with the single-embodiment grasp success metrics, with only the DEX-EE showing a slight performance decrease. These performance gaps nearly cease when further fine-tuning is performed by emphasizing the gripper through loss reweighting. The humanoid grippers show a statistically significant increase in grasp diversity over all scenes. Using principal component analysis on the synthesized grasps of the Shadow Hand, we identify significant structural differences, which indicate induced structure from joint training.
In a second experiment, we perform a direct comparison against a re-implemented
version of Diffusion~\cite{freiberg2025diffusion}. To ensure a fair comparison and
isolate the impact of the core learning objective, we use the same encoder pipeline
for both methods, holding all other architectural and training parameters constant.
We generated a new dataset for this experiment by restricting it to the static
pre-grasp configurations required by the baseline; for the Shadow Hand, we used the
three-fingered configuration from the original work~\cite{freiberg2025diffusion}.
Since it is only used for this benchmark, we reduced this dataset to only three
representative gripper types and generated 15{,}000 scenes in total. Accordingly,
our model was adapted to this setting by providing the fixed joint configurations as
static inputs. The results are presented in
Table~\ref{tab:static_gripper_comparison}.
While we do not observe a clear advantage of one method over the other in this
benchmark, diffusion-based approaches are significantly harder to deploy for the
general case where all gripper DoFs are learned through geometric representations.
Given a uniform distribution over the full action space, the flow-based approach
respects action bounds within physically plausible limits throughout the process.
Violating this requirement, as in diffusion-based approaches, results in physically
impossible joint settings and rotations exceeding 360 degrees, which the model
cannot distinguish from valid gripper settings, resulting in training instabilities.

\subsection{Real-World Validations}
We use the Franka Panda hardware platform in a grasping setting. Our method provides grasp
candidates on captured point clouds. To rank the synthesized grasp proposals for stability,
we trained an additional lightweight classifier. The classifier network relies on the same 
model architecture as our core method; however, time conditioning was removed, and 
information was
accumulated after the scene relation stage to predict the classification score.
To record the scene, we mounted a RealSense camera on the gripper and captured dense point clouds
from four different angles. To ensure a collision-free approach and pre-grasp pose, we applied an analytical,
point-cloud-based filtering algorithm. Evaluations were performed on nine novel objects in two separate
sequences, each containing five objects.

\section{Limitations and Future Work}

Addressing dexterous grasping is significantly more complex than the binary open-close problem statement presented in~\cite{freiberg2025diffusion}, necessitating a larger dataset while maintaining feasible computational costs. This led to several trade-offs in our data generation pipeline. The convex decomposition of objects required by MuJoCo~\cite{todorov2012mujoco} proved to be a bottleneck; certain objects yielded a disproportionately high number of components, making stability evaluation a computationally expensive endeavor, particularly for high-DoF grippers. To manage this, we set a hard cutoff, excluding any object that did not yield at least 5,000 stable grasps for at least one gripper. To mitigate the data generation costs and the dataset bias towards parallel-jaw grippers, we curated a representative subset of grippers. While this slightly reduces the complexity of individual scenes, it allowed us to increase the total scene count. Furthermore, our unified grasp generation strategy for dexterous grippers, which relies on a contact-matching objective, primarily produces fingertip grasps. Consequently, the diversity of grasp types (e.g., tripod, quadrupod) is dictated by the gripper's embodiment rather than a learned strategy. Addressing this limitation to generate a wider variety of grasp taxonomies is an important direction for future work.

A further limitation stems from an architectural trade-off made to simplify the model. In contrast to Diffusion~\cite{freiberg2025diffusion}, our current approach does not include an explicit geometric representation of the gripper's body during grasp synthesis. This simplification leads to an increased rate of collisions between the gripper and the scene.
The collision rate varies primarily based on gripper geometry and scene complexity, demonstrating similar trends across both single- and multi-embodiment experiments. For example, we measured median collision rates of 9.8\%, 12.5\%, and 21.0\% for the Panda, DEX-EE, and Shadow Hand grippers, respectively. While we currently filter these invalid grasps in a post-processing step, efficiently re-integrating the full gripper geometry e.g.
through use of gripper collision meshes into the model is a key priority for future work.
Our geometric joint representation can naturally accommodate under-actuation via masking during the feature readout; however, we leave the implementation of this capability for future work. As a result, joint constraints are currently enforced softly, learned implicitly from the training data. Experiments with partial point cloud
observations were omitted due to increased engineering efforts with variable sized point clouds in JAX \cite{jax2018github}.
We optimized each module using JAX's vector-mapping capabilities to enable parallelization across scenes, grippers, and grasps. As a result, our method can generate a batch of 100 grasps for the dexterous grippers in under 10 seconds on a consumer-grade GPU, a task previously infeasible with comparable Torch-based implementations~\cite{ryu2024diffusion}. Although equivariant methods have a reputation for being computationally intensive, some of our optimized modules are already highly efficient.
For instance, the Kinematic Encoder processes a batch of 128 configurations over all five grippers i.e., 640 grasps configurations total, in under 10\,ms for one training iteration. This level of performance suggests that such geometric encoders are becoming viable for latency-sensitive tasks beyond grasping. Such efforts are well-justified, as the promise of equivariant methods extends beyond mere efficiency; a strong argument for their use in manipulation is their native capacity to represent vector-based information, such as friction forces and surface normals, while seamlessly integrating scalar features from color or foundation models.

During development, it became evident that learning a universal kinematic encoder for zero-shot grasp generation was infeasible with a dataset of only five grippers. We therefore adopted a simpler approach, using fixed, learnable embeddings for each joint to reduce 
compute compared to a full encoder learning per class fixed features. We hypothesize that with a sufficiently large and diverse dataset of gripper embodiments, this encoder could be learned, enabling true zero-shot grasp synthesis for unseen grippers \cite{ai2025towards}.
\section{Conclusion}
In this work, we present an equivariant, multi-embodiment grasp synthesis method.
We demonstrate the effectiveness of using equivariant representations to encode joint kinematics, which provides a powerful geometric bias that generalizes across both prismatic and revolute joints. Our experiments empirically validate that this method achieves performance competitive with state-of-the-art techniques for single-embodiment grasp synthesis.
To facilitate future research, we release our memory- and compute-efficient JAX~\cite{jax2018github} implementation, which includes the kinematic models for five distinct grippers and the datasets used in our evaluation.

\label{sec:conclusion}
\bibliographystyle{IEEEtran}
\bibliography{reference}

\end{document}